# A Survey of AI Text-to-Image and AI Text-to-Video Generators

Aditi Singh
*Department of Computer Science*
*Kent State University*
Ohio, USA
asingh37@kent.edu

*Abstract*— **Text-to-Image and Text-to-Video AI generation models are revolutionary technologies that use deep learning and natural language processing (NLP) techniques to create images and videos from textual descriptions. This paper investigates cutting-edge approaches in the discipline of Text-to-Image and Text-to-Video AI generations. The survey provides an overview of the existing literature as well as an analysis of the approaches used in various studies. It covers data preprocessing techniques, neural network types, and evaluation metrics used in the field. In addition, the paper discusses the challenges and limitations of Text-to-Image and Text-to-Video AI generations, as well as future research directions. Overall, these models have promising potential for a wide range of applications such as video production, content creation, and digital marketing.**

## I. INTRODUCTION

With rapid advancements of deep learning and natural language processing (NLP) techniques in recent years, AI text-to-image and AI text-to-video generators have emerged as an advanced powerful tool that allows generation of images and videos from textual descriptions. These AI generators analyze the textual data using advance and complex techniques such as attention based Recurrent Neural Network [1], Generative Adversarial Network [2], and transformers [3][4] in order to generate corresponding high-quality images or videos.

The motivation after the AI text-to-image and AI text-to-video generators is driven by the necessity to automate the content creation process, making it faster for producing diverse content in proficient and economical way. These systems have potential applications in different fields such as marketing, education, and entertainment content creation. For instance, in marketing AI text-to-image generators can create product design, catalogs and user manuals [5]. In education, AI text-to-video generators can be used for creating instructional videos and animation to improve overall learning experience [6]. In entertainment industry, AI text-to-image generator and AI text-to-video generator can be used to create movie promotional videos, teasers, and more [7]. Overall, these generators aim to enhance user engagement and improve the user experience.

However, there are several limitations and challenges with the rapid development of AI text-to-image and AI text-to-video generators. The major challenge is the need of a larger high-quality training dataset. It can be challenging to obtain and label a large dataset for training. Another challenge is the lack of interpretability of the generated outputs which makes it difficult to understand the reasoning behind the generated visual content. Besides, these systems may not always be associated with the intended message or vision, leading to errors and conflicts in the generated output. Another challenge is the trade-off between visual quality and processing time. Producing high-quality images and videos can be computationally costly and slow, making it difficult to generate large amounts of content fast. Also, the produced content may not always be associated with social or public norms, leading to misinterpretation or misrepresentation of the envisioned message. These limitations and drawbacks must be carefully studied when utilizing AI text-to-image and text-to-video generators in practice.

This paper aims to provide an overview of the current state-of-the-art techniques in both AI text-to-image and text-to-video generators. Primarily to examine the underlying technologies such as data preprocessing techniques, the types of neural networks, and the evaluation metrics used in the AI text-to-image and text-to-video generators.

The paper is structured as follows. Section II provides an overview of AI text-to-image generators, including popular techniques and comparison of their capabilities. Section III explores popular AI text-to-videos generators and provide a comparison of their capabilities. Section IV present the analysis of the current state-of-the-art in both AI text-to-image and AI text-to-videos generators. Finally Section V draws conclusion.

## II. AI TEXT-TO-IMAGE GENERATOR

AI text-to-image generator are powerful tools that are equipped with natural language processing capabilities and computer vision to generate images. Let's discuss some of the popular state-of-the-art text-to-image generators such as CogView2, DALL-E 2 and Imagen. Table 1 provides details of these systems.

### A. CogView2

CogView2 [8] is an AI text-to-image generator that uses a hierarchical transformer-based approach to generate images from textual descriptions. Cogview2 uses the Cross-Modal general Language Model (CogLM), a pre-trained 6B-parameter transformer with a self-supervised task to mask and predict various types of tokens in a text and image token sequence. The hierarchical design of CogView2 allows for fast and efficient generation of high-resolution images by generating low-resolution images first and then refining them via an iterative super-resolution module that uses local parallel autoregressive generation. CogView2 is 10 times faster than CogView [9], which used sliding-window super-resolution, for generating images of similar resolution and better quality. Additionally, CogView2 supports interactive text-guided editing of images.

TABLE I. AI TEXT-TO-IMAGE GENERATORS AND THEIR COMPARISON

|  | Architecture | Training Data | Image Quality | Computational Requirements | Interpretability | Novelty |
|---|---|---|---|---|---|---|
| CogView2 | Hierarchical transformer-based | Large-scale text-only datasets | High-resolution, better quality | Relatively efficient | Difficult to interpret | Fast and efficient generation |
| DALL-E 2 | Large-scale transformer language model with StyleGAN2 architecture | Large dataset of image-text pairs | High-quality and diverse | High computational cost | Easy to interpret | Complex and diverse textual prompts |
| Imagen | Large-scale frozen T5-XXL encoder and diffusion models | Large text-only corpora | High-quality and photorealistic | Relatively efficient | Easy to interpret | Leveraging existing language models for image generation |

### B. Dall-E 2

DALL-E 2 [10] is another state-of-the-art AI text-to-image generator. It is built by OpenAI upon the success of the original DALL-E model. The main idea behind DALL-E 2 is to generate high-resolution (1024x1024) images from textual input by training a large transformer model with a 175B parameter, making it the largest language model trained to date. Unlike the original DALL-E, which used a simple VQ-VAE architecture for image generation, DALL-E 2 uses a StyleGAN2 architecture, which is a more powerful generative model that can generate more realistic and diverse images. Additionally, DALL-E 2 can handle more complex and diverse textual prompts, such as questions or instructions, and can generate a wider range of objects and scenes. To train DALL-E 2, OpenAI collected a large dataset of image-text pairs and used a multi-stage training process that combines a pretraining stage on a large text corpus with a finetuning stage on the image-text dataset. During inference, given a textual prompt, DALL-E 2 generates a sequence of image tokens in an autoregressive manner, with each token representing a patch of the final image. Finally, these image tokens are passed through the StyleGAN2 generator to produce the final high-resolution image.

DALL-E 2 has shown impressive results in generating high-quality and diverse images that are closely related to the input text. However, it still suffers from the autoregressive generation process that limits its speed and scalability, and the high computational cost of training a large-scale transformer model on a large image dataset.

### C. Imagen

Google Imagen [11] is AI text-to-image generator that combines the power of large transformer language models and the strength of diffusion models to generate high-quality image. It is built on a large frozen T5-XXL encoder, which encodes the input text into embeddings, and a conditional diffusion model that maps the text embedding into a 64x64 image.

In addition, Imagen utilizes text-conditional super-resolution diffusion models to upsample the image from 64x64 to 256x256 and from 256x256 to 1024x1024. Imagen achieves state-of-the-art results in terms of FID score and image-text alignment on the COCO dataset and outperforms recent methods in side-by-side comparisons on the comprehensive and challenging benchmark for text-to-image models, DrawBench.

Imagen discovered that large-scale language models, such as T5, can be highly efficient in encoding text for the purpose of image synthesis. This is especially true when the models have been pre-trained on text-only corpora, indicating the possibility of leveraging existing language models for image-generation tasks. This allows for a higher degree of photorealism and a deeper level of language understanding in the generated images.

As seen in above discussion, these popular AI text-to-image generators like CogView2, DALL-E 2, and Imagen use a variety of methods to produce images from text input. The hierarchical transformer-based method used by CogView2 enables the quick and effective creation of high-resolution images. DALL-E 2 is based on a substantial transformer language model and employs a potent StyleGAN2 architecture to produce a wide range of lifelike visuals. Imagen blends the strength of diffusion models with the power of massive transformer language models to create high-quality images. The ability of all three to produce various, high-quality images that are closely related to the input text has produced outstanding results.

## III. AI TEXT-TO-VIDEO GENERATOR

AI text-to-video generators have drawn a lot of interest recently because of their potential to completely change the video production sector. With the help of these generator, users may quickly and easily create highly personalized and interesting video material. These systems make use of recent developments in deep learning and natural language processing to generate films from written descriptions. While the quality and variety of videos that early AI text-to-video generators could make were constrained, newer improvements have demonstrated encouraging results in producing a wide range of extremely realistic videos. The capacity to generate videos with high levels of consistency and the requirement for large computational resources are just two of the limitations.

The following section will discuss state-of-the-art AI text-to-video generators such as Make-A-Video, Imagen Video, Phenaki, GODIVA and CogVideo highlighting their advantages, disadvantages, and prospective uses. Table 2 provides details of these models.

### A. Make-A-Video

Make-A-Video [12] is an innovative method that extends a diffusion-based text-to-image (T2I) model to text-to-video (T2V) generation through a spatiotemporally factorized diffusion model. By leveraging joint text-image priors, this approach eliminates the need for paired text-video data, enabling the potential scaling to larger amounts of video data.

For the first time, super-resolution strategies in both spatial and temporal dimensions are presented, generating high-definition, high frame-rate videos based on user provided textual input. Make-A-Video is thoroughly evaluated against existing T2V systems, showcasing state-of-the-art results in both quantitative and qualitative measures. This evaluation surpasses the existing literature in the T2V domain.

*B. Imagen Video*

Imagen Video [13] utilizes a frozen T5 text encoder, a base video diffusion model, and interleaved spatial and temporal super-resolution diffusion models to generate high quality videos. The system has been scaled up to generate 128 frame 1280x768 high-definition videos at 24 frames per second. Additionally, the system has a high degree of controllability and world knowledge, which allows it to generate diverse videos and text animations in various artistic styles, as well as with 3D object understanding. The design decisions made in the system, such as the use of fully-convolutional temporal and spatial super-resolution models and the v-parameterization of diffusion models, contribute to its successful performance

*C. Phenaki*

Phenaki [14] is another efficient and lightweight model from Google that can generate videos from short text inputs. However, it is limited to simple actions and movements and lacks fine-grained details. Phenaki is a text-to-video model that can generate long, temporally coherent and diverse videos conditioned on open-domain prompts and even sequences of prompts that tell a story.

To achieve this, Phenaki introduces a novel encoder-decoder architecture called C-ViViT, which compresses videos to discrete embeddings (tokens) and exploits temporal redundancy to improve reconstruction quality while compressing the number of video tokens. The model also uses a transformer to translate text embeddings generated by a pre-trained language model, T5X, to video tokens. Phenaki is trained on both text-to-video and text-to-image datasets and demonstrates the ability to generalize beyond what is available in the video datasets.

*D. CogVideo*

CogVideo [15] is a large-scale pretrained text-to-video generative model, which is of 9.4 billion parameters and trained on 5.4 million text-video pairs. This model utilizes the foundation provided by the pretrained text-to-image model, CogView2, effectively harnessing the knowledge acquired during the text-image pretraining phase. The model is designed to generate high-resolution (480x480) videos from natural language descriptions. To ensure the alignment between text and its temporal counterparts in the video, CogVideo uses a multi-frame-rate hierarchical training strategy. This allows the model to control the intensity of changes during the generation and significantly improves the generation accuracy, especially for movements of complex semantics.

*E. GODIVA*

GODIVA [16] is a cutting-edge text-to-video generation model that employs a Transformer architecture, pretrained on an extensive text corpus. It is capable of producing high-quality videos with increased model capacity, albeit at the expense of computational resources and extensive training data requirements. The model comprises a VQ-VAE autoencoder trained to represent continuous video pixels as discrete video tokens, and a 3D sparse attention model trained using language input and discrete video tokens as labels. This attention mechanism considers temporal, column, and row information to generate videos effectively.

GODIVA is pretrained on the HowTo100M dataset, and it demonstrates impressive video generation performance in both fine-tuning and zero-shot settings.

*F. NUWA*

NUWA [17] is a unified multimodal pre-trained model designed for visual synthesis tasks, including image and video generation and manipulation. It is a 3D transformer encoder-decoder framework that covers language, image, and video for different visual synthesis scenarios. The encoder takes text or visual sketch as input, and the decoder is shared by eight visual synthesis tasks.

To reduce computational complexity and improve the visual quality of the generated results, NUWA employs a 3D Nearby Attention (3DNA) mechanism that considers the locality characteristic of both spatial and temporal axes. 3DNA allows NUWA to efficiently process high-dimensional visual data, making it possible to scale up to larger and more complex visual synthesis tasks.

NUWA has been evaluated on eight downstream visual synthesis tasks and compared with several strong baselines, achieving state-of-the-art results in text-to-image generation, text-to-video generation, video prediction, and more. It also shows surprisingly good zero-shot capabilities on text-guided image and video manipulation tasks, meaning that it can perform these tasks without any explicit training data.

IV. Analysis

The current state-of-the-art in both AI text-to-image and text-to-video generators have demonstrated great breakthroughs. However, there exist some challenges that need to be addressed.

AI text-to-image generators have produced high-quality images with increased diversity and photorealism, and some models even offer interactive text-guided image manipulation. Unfortunately, these generators continue to necessitate a large amount of computational resources, limiting their scalability and accessibility. Moreover, existing models have depended heavily on pre-existing datasets, limiting their applicability to specific fields. Future study should emphasize improving the efficiency and availability of these generators, as well as their applicability to other areas.

AI text-to-video generators have demonstrated excellent successes in creating very realistic and customized videos, but they still face issues in generating videos with high degrees of coherence and demanding large processing resources. Recent developments generate high-quality videos using approaches such as large-scale pre-trained language models and generative models such as GANs and diffusion models, however these models are computationally costly and have scaling constraints. Future research should focus on creating new ways to enhance the production process and minimize processing expenses to make text-to-video generators more accessible and efficient.

TABLE II. AI TEXT-TO-VIDEO GENERATOR AND THEIR COMPARISON

| Video Generator | Model Type | Advantages | Limitations |
|---|---|---|---|
| Make-A-Video | 2I (Text-to-Image) models and unsupervised learning on unlabeled video data | Accelerated training, Unsupervised learning, Inheritance of image generation models. | Cannot learn associations between text and certain phenomena in videos. |
| Imagen Video | cascade of video diffusion models | High fidelity, Diverse video generation, 3D object understanding, Text animations | Trained on problematic data [18][19][20], social biases, and stereotypes. |
| Phenaki | Encoder-decoder model with a transformer | Good performance on video prediction, can generate long videos conditioned on text and starting frame | Trained on biased datasets. |
| GODIVA | Text-to-video pretrained model with three-dimensional sparse attention mechanism | Reduced computation cost, Good zero-shot capability | Challenge to generate long videos with high resolution, evaluating text-to-video generation task remains a challenge. |
| CogVideo | Inherits pretrained text-to-image model CogView2 | Efficiently leverages image generation capacity, better understanding of text-video relations | Restriction on input sequence length, large scale model and limitation of GPU memory. |
| NUWA | Multimodal pretrained model with 3D transformer encoder-decoder framework | Reduces computational complexity, Good zero-shot capabilities | Poor text-video alignment in frames. |

## V. CONCLUSION

In this paper, a brief description of different types of AI text-to-image and AI text-to-video generators has been presented. The future of AI text-to-image and AI text-to-video generators appears bright. Continued research and development in these areas are likely to result in more efficient, powerful, and accessible systems that can revolutionize the way users produce and interact with digital content. AI text-to-image and AI text-to-video generators have the potential to usher in a new era of creativity and productivity in a variety of industries due to their ability to create high-quality images and videos from textual descriptions. As such, they will undoubtedly stay an active area of research and development in the coming years.